%% file: main.tex
\documentclass[lettersize,journal]{IEEEtran}
\usepackage{amsmath,amsfonts}
\usepackage{algorithmic}
\usepackage{algorithm}
\usepackage{array}
\usepackage[caption=false,font=normalsize,labelfont=sf,textfont=sf]{subfig}
\usepackage{textcomp}
\usepackage{stfloats}
\usepackage{url}
\usepackage{verbatim}
\usepackage{graphicx}
\usepackage{cite}
\usepackage{etoolbox}

\usepackage{bm}
\usepackage{amsmath}
\usepackage{amsfonts}
\usepackage{makecell}
\usepackage{CJKutf8}
\usepackage[utf8]{inputenc}

\usepackage{xspace}

\usepackage[numbers]{natbib}
\usepackage{booktabs}

\usepackage{graphicx}
\usepackage{caption}

\newcommand{\eg}{e.g.\xspace}
\newcommand{\ie}{i.e.\xspace}

\begin{document}

\title{Large Material Gaussian Model for Relightable 3D Generation}

\author{Jingrui Ye$^{*}$, Lingting Zhu$^{*}$, Runze Zhang, Zeyu Hu, Yingda Yin, Lanjiong Li, Lequan Yu$^{\dag}$, Qingmin Liao$^{\dag}$
        % <-this % stops a space

\thanks{ J. Ye and Q. Liao are with Tsinghua Shenzhen International Graduate School.
L. Zhu and L. Yu are with The University of Hong Kong. 
R. Zhang, Z. Hu, and Y. Yin are with LIGHTSPEED.
L. Li is with The Hong Kong University of Science and Technology (Guangzhou). $^{*}$Equal contribution. $^{\dag}$Corresponding authors.
}
}

\markboth{Journal of \LaTeX\ Class Files,~Vol.~14, No.~8, August~2021}%
{Shell \MakeLowercase{\textit{et al.}}: A Sample Article Using IEEEtran.cls for IEEE Journals}



\maketitle

\begin{abstract}
The increasing demand for 3D assets across various industries necessitates efficient and automated methods for 3D content creation. Leveraging 3D Gaussian Splatting, recent large reconstruction models (LRMs) have demonstrated the ability to efficiently achieve high-quality 3D rendering by integrating multiview diffusion for generation and scalable transformers for reconstruction. However, existing models fail to produce the material properties of assets, which is crucial for realistic rendering in diverse lighting environments. In this paper, we introduce the \textbf{Large Material Gaussian Model (MGM)}, a novel framework designed to generate high-quality 3D content with Physically Based Rendering (PBR) materials, \ie, albedo, roughness, and metallic properties, rather than merely producing RGB textures with uncontrolled light baking. Specifically, we first fine-tune a new multiview material diffusion model conditioned on input depth and normal maps. Utilizing the generated multiview PBR images, we explore a Gaussian material representation that not only aligns with 2D Gaussian Splatting but also models each channel of the PBR materials. The reconstructed point clouds can then be rendered to acquire PBR attributes, enabling dynamic relighting by applying various ambient light maps. Extensive experiments demonstrate that the materials produced by our method not only exhibit greater visual appeal compared to baseline methods but also enhance material modeling, thereby enabling practical downstream rendering applications.
\end{abstract}


\input{sections/1-intro}

\input{sections/2-related}

\input{sections/3-preliminary}
\input{sections/4-methods}
\input{sections/5-experiments}

\section{Limitations} 
Although our method successfully generates diverse and high-quality 3D assets with PBR material according to the text prompts, it still has several limitations. First, since material estimation is an inherently ill-posed problem, our multiview PBR diffusion exhibits inaccurate roughness and metallic generation under some circumstances. Second, our method struggles with materials that exhibit properties like transparency, high reflection, or subsurface scattering. This limitation arises primarily from our choice of the Bidirectional Reflectance Distribution Function (BRDF) model, which is not equipped to handle more advanced and complex materials. 
Furthermore, for the reconstructed model, our approach faces challenges in recovering high-frequency geometry and texture details, mainly due to the limitation of Gaussian volume representation based reconstruction model.

\section{Conclusion}

In this works, we pioneeringly propose a large multiview Gaussian model for generating high-quality 3D assets with PBR material, and support relighting with different environment maps. Distinct from previous methods can only generate shaded RGB texture, we purpose a novel \textbf{material Gaussian} representation together with the well designed training strategy to learn the material efficiently.
Additionally, we inject the \textbf{geometry prior} into both our generation and reconstruction stages to better supervise the material reconstruction, improving the completeness and detail of generated geometry and appearance.
In summary, we presents the first generative methods for producing Gaussians with PBR and achieves state-of-the-art text-to-3D generation quality based on Gaussian Splatting representation, the versatility and applicability have been proved across various contexts.

\bibliographystyle{IEEEtran}
% \small
\footnotesize
% \scriptsize
\bibliography{acmart}

\vfill

\end{document}

%% file: sections/1-intro.tex
\section{Introduction}
Automatic 3D content creation holds immense potential across various domains, including digital gaming, virtual reality, and filmmaking.
% Traditional approaches to generating 3D assets have utilized optimization-based techniques from multi-view posed images~\cite{yariv2021volume, wang2021neus}, 
Recently, 3D generative models achieve significant advancements, and they have remarkably reduced huge manual labor of professional 3D artists and democratized creativity in 3D asset generation for non-experts.
Previous optimization-based methods~\cite{poole2022dreamfusion, wang2024prolificdreamer, lin2023magic3d, chen2024vividdreamer, liang2024luciddreamer, qiu2024richdreamer} in 3D generation have focused predominantly on score distillation sampling (SDS) with 2D diffusion model priors, which is often hindered by expensive optimization, the Janus problem, and inconsistency between different views. 
Recent advancements have significantly shortened inference time by leveraging large reconstruction model (LRM), which usually take multiview images as input. Some approaches~\cite{hong2023lrm, li2023instant3d, wei2024meshlrm, tang2025lgm, xu2024grm, xu2024instantmesh, xu2023dmv3d, chen2025lara} utilize transformers~\cite{vaswani2017attention} to learn the triplane-based~\cite{chan2022efficient} neural radiance fields (NeRF)~\cite{mildenhall2021nerf}, but fail to reconstruct detailed geometry and texture due to the low-resolution nature and inferior post-conversion. 
More recently, several works~\cite{tang2025lgm, he2025gvgen, chen2025lara, xu2024grm, wang2025crm, chen2024generalizable} employ 3D Gaussian Splatting~\cite{kerbl20233d} or its variant~\cite{huang20242d} as 3D representation and facilitate high-resolution training with expressiveness, combined with attention mechanism to generate Gaussians from multiview images.

Despite these advancements, such approaches still struggle to control light shading and do not account for the material properties of the 3D objects, \eg, Physically Based Rendering (PBR) materials. 
Consequently, the reconstructed 3D objects often exhibit uncontrolled light baking in textures, resulting in a lack of realistic rendering effects and an inability to adapt appearances under varying ambient lighting conditions. This significantly limits their applicability in downstream applications.
To address these limitations, as far as we know, we are the first to propose a novel method for generating 3D Gaussians with PBR materials, which are adaptable for relighting during Gaussians rendering. Our approach consists of two main components: a generative model and a reconstruction model, starting from a text prompt and finally producing 3D rendering incorporating PBR. Initially, we train a text-to-PBR multiview diffusion model on 3D material datasets, built upon multiview diffusion models, \eg, MVDream~\cite{shi2023mvdream}. Subsequently, a large reconstruction model based on 2D Gaussian Splatting~\cite{huang20242d} is employed to reconstruct the sparse views of the generated PBR images following the designs and architectures of recent Gaussian-based reconstruction models~\cite{tang2025lgm, chen2025lara}. To ensure global consistency across both the generation and reconstruction stages, we incorporate geometry data, \ie, depth and normal maps, to serve as proxy information. The integration offers benefits in three folds: 1) We can seamlessly adopt geometry generation models, \eg, ~\cite{li2024craftsman,tochilkin2024triposr,wu2024unique3d,xu2024instantmesh, chen2025mar} for additional geometry guidance and mitigate multiview inconsistency; 2) The assistance of geometry guidance models, \eg, text-to-geometry models and text-to-depth/normal models~\cite{qiu2024richdreamer}, can be optionally removed when untextured 3D models are available to produce rendering geometry maps, aligning more closely with practical usages; 3) Since the PBR images inherently lack lighting cues and differ significantly from the distribution of shaded images, relying solely on them for training radiance fields produces Gaussian points with inaccurate spatial details, indicating the geometry guidance is necessary in reconstruction. In practice, we design a geometry-conditioned multiview generation model with ControlNet~\cite{zhang2023adding} in the generation stage, and then introduce geometry features injection and utilize geometric guidance to produce additional supervision in reconstruction. 
Incorporating efficient rasterization, our method achieves high-resolution renderings along with albedo, roughness, and metallic maps, enabling dynamic relighting effects under various ambient lighting conditions.

Extensive qualitative and quantitative experiments validate the effectiveness of our method, demonstrating that it matches or exceeds state-of-the-art Gaussian Splatting based generation quality while achieving realistic rendering results under diverse lighting conditions. Our contributions are summarized as follows:
% \vspace{-0.5em}
\begin{enumerate}
    \item We pioneer to propose a novel framework to generate high-resolution Gaussians with material properties from text prompts, capable of being dynamically relighted through physically based rendering.
    \item We introduce principled models tailored to relightable Gaussians, including a controllable multiview PBR model and a unified material Gaussian reconstruction model.
    \item Extensive experiments demonstrate that our method excels in both appearance and material quality with efficiency.
\end{enumerate}

%% file: sections/2-related.tex
\section{Related Work}
\subsection{Large Reconstruction Model}
Recent advancements in NeRF~\cite{mildenhall2021nerf} and 3DGS~\cite{kerbl20233d} enable high-quality novel view synthesis % and generation capabilities. 
capabilities. Pioneered by the large reconstruction model (LRM)~\cite{hong2023lrm}, recent works~\cite{wei2024meshlrm, li2023instant3d, xu2023dmv3d, xu2024instantmesh} demonstrate that image tokens can be directly mapped to 3D representations, typically triplane-NeRF, in a feed-forward manner via a scalable transformer-based architecture~\cite{vaswani2017attention} with large-scale 3D training data~\cite{wu2023omniobject3d, deitke2023objaverse, yu2023mvimgnet}. Among them, Instant3D~\cite{li2023instant3d} integrates LRM with multiview diffusion models~\cite{wang2023imagedream, shi2023mvdream, shi2023zero123++, liu2023zero, lu2024direct2}, using four views of generated images for better quality. To avoid inefficient volume rendering and limited triplane resolution, some concurrent works~\cite{xu2024grm, tang2025lgm, zhang2025gs} follow Instant3D and introduce 3DGS or its variants~\cite{huang20242d} into sparse-view LRM variants for more efficient rendering. Specifically, LGM~\cite{tang2025lgm} combines 3D Gaussians from different views using a convolution-based asymmetric U-Net~\cite{ronneberger2015u}, along with other Gaussian reconstruction models like~\cite{wang2025crm, he2025gvgen, zou2024triplane, zhang2024geolrm}. GRM~\cite{xu2024grm} and GS-LRM~\cite{zhang2025gs} use pixel-aligned Gaussian~\cite{szymanowicz2024splatter, charatan2024pixelsplat} with a pure transformer-based reconstruction model.
LaRa~\cite{chen2025lara} adopts 2DGS~\cite{huang20242d} representations and models scenes as Gaussian volumes and group attention layers for better quality and faster convergence.

\subsection{Optimization-based 3D Generation}
In 3D generation, due to the scarcity of 3D data, leveraging 2D priors has become an explorable method.
DreamFusion~\cite{poole2022dreamfusion} pioneers the optimization of 3D assets by distilling from pre-trained image diffusion models~\cite{rombach2022high}, followed by a large group of successors~\cite{wang2023score, lin2023magic3d, wang2024prolificdreamer, sun2023dreamcraft3d} with more advanced distillation techniques. 
Some methods explore more efficient 3D representations, \eg, hashgrid~\cite{muller2022instant} and 3DGS~\cite{kerbl20233d}, to accelerate the optimization~\cite{lorraine2023att3d, tang2023dreamgaussian, yi2023gaussiandreamer, chen2024text}. DreamGaussian proposes a two-stage coarse-to-fine Gaussian generation method, and performs texture refinement in UV space. GaussianDreamer~\cite{yi2023gaussiandreamer} combines Gaussian distribution with 3D and 2D diffusion models~\cite{rombach2022high, jun2023shap}, 
and introduces noise point growth and color perturbation to enrich the details. 
However, these optimization processes still incur additional computational costs at test time. Moreover, since 2D diffusion models inherently lack multiview awareness, these methods frequently struggle with Janus problem. And approaches based on SDS loss tend to produce over-saturated colors, leading to suboptimal generation results.

\begin{figure*}[h]
    \centering
    \includegraphics[width=0.93\textwidth]{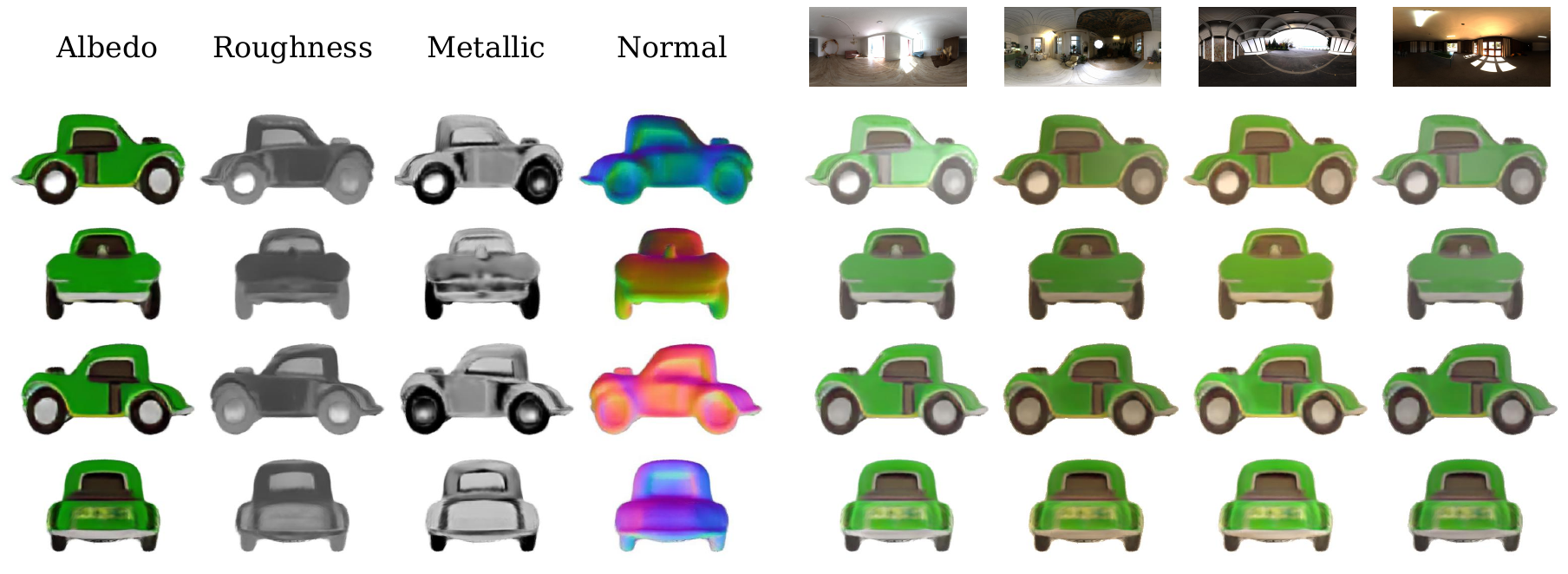}
    % \vspace{-1em}
    \captionof{figure}{The text prompt of the above case is \emph{``a green single-door toy car''}. Taking the textual description as input, our method can generate high-quality Gaussians consisting of \textbf{albedo, roughness, and metallic} that can be \textbf{relighted} for photo-realistic rendering under any new illumination environments. We invite readers to read our appendix for more relighting results.}
    \label{fig:teaser}
\end{figure*}

\subsection{PBR Material and Relighting}
3D objects include both geometric and material properties. Predicting only the radiance has a significant limitation: it fails to produce convincing results when relighting, as material properties such as reflectance and roughness are not captured. Therefore, a stream of researches~\cite{liu2025unidream,qiu2024richdreamer,chen2023fantasia3d,wang2024boosting,li2024idarb,shim2024mvlight} focus on material generation and downstream relighting,
Fantasia3D~\cite{chen2023fantasia3d} introduces the bidirectional reflectance distribution function (BRDF)~\cite{cook1982reflectance} to the task of text-to-3D, and optimizes PBR materials with SDS loss. By fine-tuning multiview diffusion, attempts like UniDream~\cite{liu2025unidream} and RichDreamer~\cite{qiu2024richdreamer} adopt an albedo-normal aligned multiview diffusion and progressive generation for geometry and albedo-textures based on SDS, which achieves multiview consistency, but suffer from high computational costs.

Some works aim at producing PBR maps for untextured meshes. Some of them~\cite{xiong2024texgaussian,wang2024boosting,huang2024material,zhang2024clay} generate single-view or multiview PBR image by fine-tuning diffusion model, and continuously update the UV map with RePaint~\cite{lugmayr2022repaint} and UV projection. Another type of methods directly trains a diffusion model~\cite{yu2023texture,yu2024texgen} or uses 2D diffusion priors~\cite{youwang2024paint} in the UV space to model PBR, improving global consistency but may suffer from poor generalization.

Other works~\cite{zhang2024relitlrm, feng2024arm, shim2024mvlight, he2024diffrelight, chen2024rgm, he2024neural, zhu2025muma} train models on multiview data with varying lighting conditions and viewpoints. ReLitLRM~\cite{zhang2024relitlrm} proposes a geometry-based transformer and adopts a diffusion-based relightable appearance generator, where the geometry is separated from the appearance to better handle the uncertainty caused by lighting. ARM~\cite{feng2024arm} decouples geometry from appearance, processes appearance within the UV space, and introduces a material prior that encodes semantic appearance information to address the material and illumination ambiguities present in sparse-view input images. However, these methods do not model material properties of objects and often require high training costs.

%% file: sections/3-preliminary.tex
\section{Preliminary}
 
\subsection{2D Gaussian Splatting}
\label{preliminary_2dgs}
3DGS~\citep{kerbl20233d} achieves great success in 3D rendering with radiance field. Yet, 
It models angular radiance as a single blob, limiting high-quality surface reconstruction.
2DGS~\cite{huang20242d} further takes advantage of standard surfel modeling~\cite{pfister2000surfels, yifan2019differentiable}, adopting 2D oriented disks as surface elements and better align with thin surfaces. 
Specifically, 2DGS evaluates Gaussian values at 2D disks and utilizes explicit ray-splat intersection, resulting in a perspective-correct splatting:
\begin{equation}
\mathcal{G}(\bm{u}) = \exp\left(-\frac{u(\bm{r})^2+v(\bm{r})^2}{2}\right),
\label{eq:gaussian-2d}
\end{equation}
where $\bm{u}=(u(\bm{r}),v(\bm{r}))$ is the intersection point between ray $\bm{r}$ and the primitive in UV space. And each Gaussian primitive has its own view-dependent color $\bm{c}$, defined by SH coefficients $\bm{f}$ with degree $\bm{k}$. 
In summary, each 2D Gaussian can be characterized as:
\begin{equation}
{\Theta} = \{\bm{x}, \bm{s}, \bm{q}, \alpha, \bm{f}\},
\label{eq:material gaussian}
\end{equation}
with position $\bm{x}\in \mathbb{R} ^{3}$, scaling vector $\bm{s}\in \mathbb{R} ^{2}$, rotation vector $\bm{q}\in \mathbb{R} ^{4}$, opacity $\alpha\in \mathbb{R} ^{1}$ and SH coefficients $\bm{f}\in \mathbb{R} ^{3\times (k+1)^{2}}$. 
For rendering, Gaussians are sorted according to their centers and composed into pixels using front-to-back alpha blending:
\begin{equation}
\label{2dgs}
\mathbf{c}(\bm{r}) = \sum_{i=1} \bm{c}_i \alpha_i \mathcal{\hat {G}}_i(\bm {u}) T_i,
\end{equation}
where $T_i$ is the approximated accumulated transmittances defined by $\prod_{j=1}^{i-1} (1 - \alpha_j \mathcal{\hat {G} }_j(\bm{u}))$ where $\mathcal{\hat{G}}$ denotes the results after a low-pass filter, and the integration process is terminated when the accumulated opacity reaches saturation. 

\subsection{Large Gaussian Models}
\label{preliminary_gaussian_model}
Leveraging the efficient rasterization of 3DGS~\cite{kerbl20233d}, recent methods such as LGM~\cite{tang2025lgm} and LaRa~\cite{chen2025lara} introduce Gaussian-based reconstruction models. 
They follow the LRM approach~\cite{hong2023lrm}, 
demonstrating significant advancements in 3D content reconstruction and creation.
Our material Gaussian reconstruction model is built upon LaRa~\cite{chen2025lara}, 
which takes $M$ images $\bm{I} \!=\! (\bm I_1, \ldots, \bm I_M)$ with camera parameters $(\bm \pi_1, \ldots, \bm \pi_M)$ to reconstruct radiance fields as a collection of 2D Gaussians.
The model operates on a 3D voxel grid and consists of three volumes: an \textit{image feature volume} $\bm{V}_\text{f}$ coping with image conditions, an \textit{embedding volume} $\bm{V}_\text{e}$ describing prior learned from data, and a \textit{Gaussian volume} $\bm{V}_\text{g}$ representing the radiance field. 

Given multiview images, feature maps of $M$ views are extracted by DINO~\cite{caron2021emerging} encoder with Plücker ray directions, then lifted to 3D to form the \textit{image feature volume} $\bm{V}_\text{f}$. A volume transformer containing a set of \emph{group attention layers} is used to predict the Gaussian volumes and each group attention layer contains three sublayers: group cross-attention, a MLP, and 3D convolution. The image feature volume $\bm{V}_\text{f}$ and embedding volume $\bm{V}_\text{e}$ are unfolded into $G$ groups along each axis, and cross-attention is applied between the corresponding groups of embedding tokens $\bm{V}_\text{e}^{g,j}$ and feature tokens $\bm{V}^{g}_\text{f}$, where $j$ denotes the index of the layer starting from 1 and $\{\bm{V}_\text{e}^{g,1}\}_g \!=\! \bm{V}_\text{e}$. The embedding groups are then updated using an MLP to produce $\{ \ddot{\bm{V}}_\text{e}^{g,j} \}^G_{g=1}$ which are reassembled into $\ddot{\bm{V}}_\text{e}^{j}$ and fed into a 3D convolutional layer. The process is described as:
\begin{align}
\dot{\bm{V}}_\text{e}^{g,j} &= \text{GroupCrossAttn}\left(\text{LN}\left(\bm{V}_\text{e}^{g,j}\right), \bm{V}^{g}_\text{f}\right) + \bm{V}_\text{e}^{g,j} \text{,} \\
\ddot{\bm{V}}_\text{e}^{g,j} &= \text{MLP}\left(\text{LN}\left(\dot{\bm{V}}_\text{e}^{g,j}\right)\right) + \dot{\bm{V}}_\text{e}^{g,j} \text{,} \\
\bm{V}_\text{e}^{j+1} &= \text{3DCNN}\left(\text{LN}\left(\ddot{\bm{V}}_\text{e}^{j}\right)\right) + \ddot{\bm{V}}_\text{e}^{j}.
\end{align}
After passing through all group attention layers, LaRa employs a 3D transposed CNN to upscale the updated embedding volume $\dot{\bm{V}}_\text{e}$ and obtain the Gaussian volume: $\bm{V}_\text{g} = \text{Transpose-3DCNN}\left(\dot{\bm{V}}_\text{e}\right)$. Finally, 2D Gaussian primitives are decoded from the Gaussian volume using a coarse-to-fine strategy and achieve efficient high-resolution image rendering through the rasterization in 2DGS~\cite{huang20242d}.

%% file: sections/4-methods.tex
\begin{figure*}[t!]
    \centering
    \includegraphics[width=\textwidth]{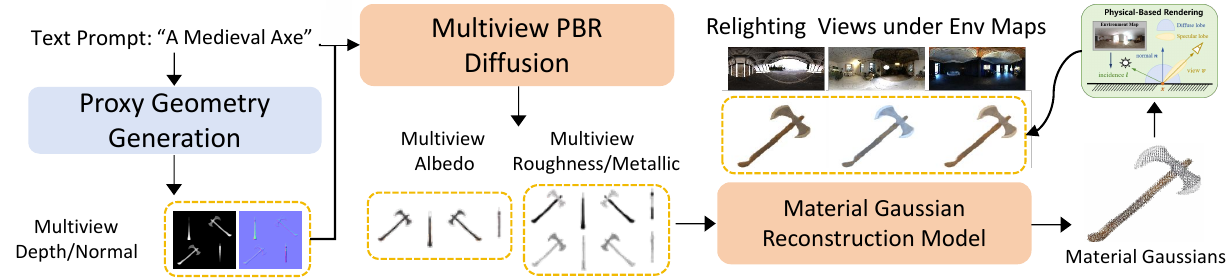}
    % \vspace{0.1mm}
    \caption{
    \textbf{Overview of our method}. Starting with a text input, our MGM generates multiview depth and normal maps, serving as geometry priors. These maps, along with the text prompt, guide the generation of multiview PBR images through our specialized multiview PBR diffusion process. Subsequently, these images inform our material Gaussian reconstruction model, which reconstructs the material Gaussians complete with PBR components. The integration of PBR materials ensures that these Gaussians can be seamlessly utilized to achieve realistic lighting effects in various rendering scenarios.
    }
    % \vspace{-0.2cm}
    \label{fig:pipeline_full_process}
\end{figure*}

\section{Methodology}
In this section, we discuss the full pipeline of our Large Material Gaussian
Model (MGM). To begin with, we introduce the designs of multiview PBR diffusion in Section~\ref{sec:Multiview PBR Diffusion}. Then, we discuss our material Gaussian reconstruction model in Section~\ref{sec:Material Gaussian Reconstruction Model}. Finally, we present that with our material Gaussian representation, downstream relighting can be seamlessly integrated in Section~\ref{sec:Relightable Gaussian}. An overview of MGM is shown in Figure~\ref{fig:pipeline_full_process}.

\subsection{Multiview PBR Diffusion}
\label{sec:Multiview PBR Diffusion}
We aim to synthesize PBR materials, including albedo of 3 channels, roughness and metallic of 1 channel each. To achieve this, we modify base model that is originally designed for shaded image space generation and fails to decompose lighting effects to suit the need of material attributes with additional channels and modalities. To re-target the base Stable Diffusion~\cite{rombach2022high} or multiview models like MVDream~\cite{shi2023mvdream} for downstream applications, the implementations have been relatively standard in recent years, \ie, preparation of downstream data and re-purposing the base models for new targets with optionally new modules, demonstrated by a wide range of applications~\cite{ke2024repurposing,fu2025geowizard,zhang2024clay,liu2023hyperhuman,qiu2024richdreamer}. Specifically, we train two sub-models where one for albedo and one for roughness/metallic. For the latter, we inflate the outermost blocks in the input and output blocks, adopting parameter copy for initialization of additional branches. We leverage the 3D datasets~\cite{deitke2023objaverse} to render a consistent multiview PBR image dataset for fine-tuning, denoted as $\mathcal{X}_{mv-PBR}$. Following the same training regimen of multiview diffusion models, our model can synthesize texture images from four viewpoints, and align precisely with the input geometry with multiple ControlNet branches~\cite{zhang2023adding} that take each target view’s rendered depth map $\bm{I_{d}}$ and normal map $\bm{I_{n}}$ as input control images $\bm I_{c}=\left \{\bm {I_{d}},\bm {I_{n}} \right \} $.

Formally, given a set of noisy image $\bm {x}_t \in \mathbb{R}^{F\times H \times W \times C}$, the feature of text prompt $\bm{y}$, and a set of extrinsic camera parameters $\bm c \in \mathbb{R}^{F\times16}$, where $F$ is the view number dimension, our multiview PBR diffusion is trained to generate a set of images $\bm{x}_0 \in \mathbb{R}^{F\times H \times W \times C}$ of the same scene from $F$ different view angles with $C\!=\!5$ channels. For training samples $\{\bm{x}, \bm{y}, \bm I_{c}, \bm{\mathbf{c}\}} \in \mathcal{X}_{mv-PBR}$,
the loss is defined as:
\begin{align}
\label{eq:cond_ldm_loss}
\mathcal{L}(\bm \theta, \mathcal{X}_{mv-PBR}) = \mathbb{E}_{\bm x, \bm y, \bm c, t, \epsilon}\Big[
\Vert \bm \epsilon - \bm {\epsilon_\theta}(\bm{x}_{t};\bm y, \bm I_{c},\bm c,t) \Vert_{2}^{2}
\Big],
\end{align}

where $\bm {x}_{t}$ is the noisy image diffused with random noise $\bm \epsilon$, and the $\bm {\epsilon_\theta}$ is the output of multiview material diffusion model $\bm \theta$. 

During inference, we use text prompts along with the geometry information to generate multiview PBR diffusion model. To obtain the multiview depth and normal conditions, we can resort to pre-trained 3D geometry generation models for supplying the proxy information, denoted as multiview depth and normal generation block shown in Figure~\ref{fig:pipeline_full_process}. In practice, we use CraftsMan~\cite{li2024craftsman} to get a 3D mesh and render the multiview images at desired viewpoints in inference. Though existing geometry generation models fail to produce high-quality untextured meshes that are really close to the distribution of the high-quality 3D assets in the datasets, thanks to the generalization capabilities of our trained multiview PBR diffusion, our method can still cope with generated multiview geometry conditions. Furthermore, as sometimes untextured 3D meshes are already available in real-world pipelines, we can directly render it to acquire geometry guidance for PBR diffusion models, indicating the wide use-cases of our model.

\subsection{Material Gaussian Reconstruction Model}
\label{sec:Material Gaussian Reconstruction Model}

A naive approach to acquire Gaussians set with multiple attributes is to train three reconstruction models for PBR material channels respectively, but this would reconstruct three point clouds with different geometry, leading to obvious artifacts in the PBR rendering and a lack of practical methods for their integration. To address this, we designed a material Gaussian representation that contains multi-channel material properties. We keep all geometric and position-related Gaussian parameters the same as Gaussian Splatting, and re-target the SH coefficients to degraded view-independent colors comprising of 5 channels, where the first 3 channels correspond to albedo, and the last 2 channels represent roughness and metallic. Here we consider that albedo does not contain any lighting components, so orders higher than 1 in SH coefficients should not be used. In summary, our material Gaussian can be represented as:
\begin{equation}
{\Theta} = \{\bm{x}, \bm{s}, \bm{r}, \alpha, \bm{f}_a, \bm{f}_r, \bm{f}_m\},
\label{eq:material gaussian}
\end{equation}
where $\bm{f}_{a}\in \mathbb{R}^{3}$ and $\bm{f}_{r/m}\in \mathbb{R}^{1}$. This unified representation aligns with the design of 2DGS but also models materials and enables training a unified model operating on PBR-aware Gaussians.

\begin{figure*}[t]
    \centering
    \includegraphics[width=0.9\textwidth]{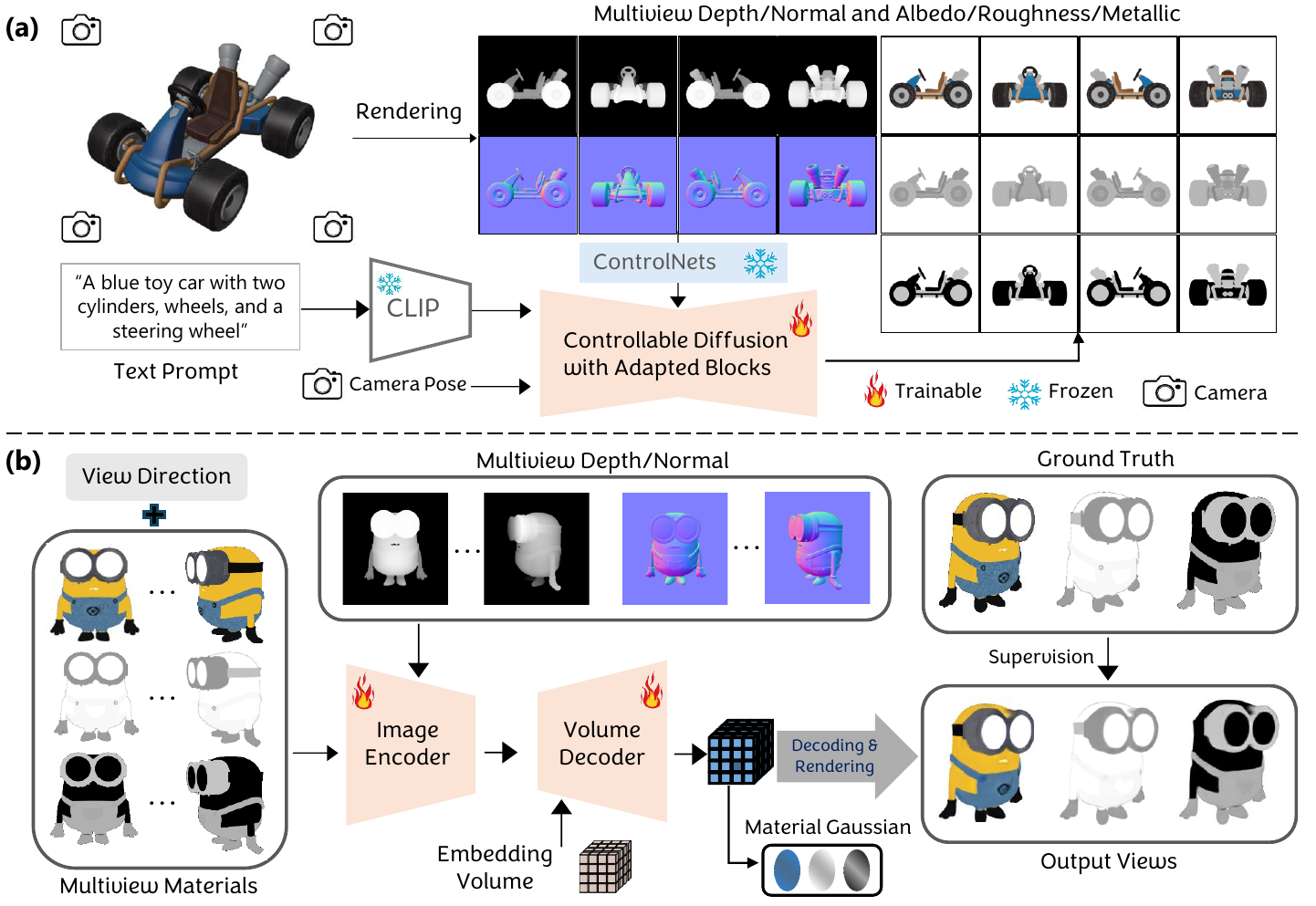}
    % \vspace{-1em}
    \caption{
    \textbf{Detailed designs of generative modules and reconstruction modules}. 
    \textbf{a) Multiview PBR Diffusion}. With the input text prompt and camera embeddings, and the rendered geometry views the controllable diffusion with adapted blocks aim at synthesizing multiview materials. \textbf{b) Material Gaussian Reconstruction Model.} Multiview images are passed through the Image Encoder to extract features with view direction injected and the Volume Decoder predicts the Gaussian volume.
    }
    \label{fig:models}
\end{figure*}

Since PBR images lack lighting clues compared to shaded images, we find that the geometry reconstruction performance when directly using PBR data for model training is inferior to that achieved with shaded data, this trend occurs for both the unified model and the three separate models. In addition, it is essential that the reconstructed geometry aligns with the proxy geometry provided to the multiview PBR diffusion model.
To address the problem above, we take depth and normal maps for the second use in the full pipeline, incorporating them into the reconstruction model to enhance the model's perception of the spatial location, redesigning the reconstruction model and adopting additionally geometry supervision. As a result, the predicted positions and opacities of the Gaussians can be more accurate under the guidance. To be specific, given $5\times N$ images containing albedo $\bm{I}_{a} \!=\! ({\bm I}_{a}^{1} , \ldots, {\bm I}_{a}^{N} )$, roughness $\bm{I}_{r} \!=\! ({\bm I}_{r}^{1} , \ldots, {\bm I}_{r}^{N} )$, metallic $\bm{I}_{m} \!=\! (\bm I_{m}^{1} , \ldots, \bm I_{m}^{N} )$, depth $\bm{I}_{d} \!=\! (\bm I_{d}^{1} , \ldots, \bm I_{d}^{N} )$, and normal $\bm{I}_{n} \!=\! (\bm I_{n}^{1}, \ldots, \bm I_{n}^{N})$; along with $N$ camera embedding $\bm{c} \!=\! (\bm c^{1} , \ldots, \bm c^{N})$ and $\bm c^{i}=(\bm{d}^i, (\bm{o}^i \times \bm{d}^i)) \in \mathbb{R} ^{6}$, where $\bm{o}$ is the camera location and $\bm{d}$ is the view direction vector. We first apply the image encoder~\cite{zhang2022dino} to extract features and concatenate with camera embedding for each type of image:
\begin{align}
h_{a}=\text{Concat}(\mathcal{F}_{a}(\bm{I}_{a}), \mathcal{F}_{g}(\bm{I}_{d}), \mathcal{F}_{g}(\bm{I}_{n}), \bm{c})\text{, } \\
h_{r/m}=\text{Concat}(\mathcal{F}_{r/m}(\bm{I}_{r/m}), \mathcal{F}_{g}(\bm{I}_{d}), \mathcal{F}_{g}(\bm{I}_{n}), \bm{c}),
\end{align}
where $\mathcal{F}_{a}, \mathcal{F}_{g}, \mathcal{F}_{r/m}$ are the image encoders for albedo, depth/normal, and roughness/metallic, respectively. With these image feature $\mathcal{H}_{f}=(h_{a}, h_{r}, h_{m})$ and an embedding volume $\mathcal{H}_{e}$ which is a learnable query embedding that aims to capture data prior dynamically, we employ a volume decoder $\Phi_{g}$ based on transformer from LaRa~\cite{chen2025lara} to predict the 3D Gaussian volume, and the input is unfolded:
\begin{align}
\mathcal{H}_{g}=\Phi_{g}(\text{Flatten}(\mathcal{H}_{f}), \text{Flatten}(\mathcal{H}_{e})).
\end{align}
Finally, our material Gaussian primitives ${\Theta}$ can be derived from $\mathcal{H}_{g}$ through a lightweight parameter decoding network. Please refer to the Appendix for more details.

\noindent \textbf{Reconstruction Loss.} 
Since our material Gaussian representation comprises of multiple channels, it is crucial to optimize the PBR components jointly to achieve a satisfactory appearance and geometry. We adopt the standard reconstruction loss in differentiable rendering to guide the appearance, utilizing 2D Gaussian Splatting~\cite{huang20242d}. At each training iteration, we render the albedo $\hat{\bm{I}}_{a}$, roughness $\hat{\bm{I}}_{r} $ and metallic $\hat{\bm{I}}_{m} $ from 8 views, including 4 reconstructed input views and 4 novel views. Following other works in LRM series~\cite{hong2023lrm, chen2025lara, tang2025lgm, xu2024grm}, we apply a straightforward image reconstruction loss between the PBR renderings from Gaussians, \ie, $\hat{\bm{I}}=(\hat{\bm{I}}_{a}, \hat{\bm{I}}_{r}, \hat{\bm{I}}_{m} ) $ and ground-truth images, \ie, ${\bm{I}}$:
\begin{align}
\mathcal{L}_{\text{Image}}=\mathcal{L}_\text{MSE}(\hat{\bm{I}}, \bm{I})+\mathcal{L}_\text{SSIM}(\hat{\bm{I}}, \bm{I}),
\label{eq:image_loss}
\end{align}
where $\mathcal{L}_\text{MSE}$ is the pixel-wise L2 loss, and $\mathcal{L}_\text{SSIM}$ is the structural similarity loss.

\noindent \textbf{Geometry Regularization.} 
The geometric priors, \ie, ${\bm{I}_{g}}=(\bm I_{d}, \bm I_{n})$, are used to extract image feature. This not only ensures that the 2D Gaussians can be optimized to produce better geometry, but also helps to maintain the consistency with the PBR generation stage, as initially controlled by the proxy geometry generation. We render the depth and normal maps of 4 input views from 2D Gaussians, \ie, ${\hat{\bm{I}}_{g}}=(\hat{\bm I}_{d}, \hat{\bm I}_{n})$ and calculate regularization loss:
\begin{align}
\mathcal{L}_{\text{Geometry}}=\mathcal{L}_\text{MSE}(\hat{\bm{I}}_{d}, \bm{I}_{d}) + \mathcal{L}_\text{MSE}(\hat{\bm{I}}_{n}, \bm{I}_{n}).
\label{eq:geometry_loss}
\end{align}

\noindent \textbf{2DGS Regularization.} 
Moreover, we add two regularization terms following 2DGS~\cite{huang20242d}: depth distortion and normal consistency. Specifically, for a ray $\mathbf{u}(\mathbf{x})$ originating from pixel $\mathbf{x}$, the weight distribution is concentrated by minimizing the distance between ray-primitive intersections, leading to the distortion loss: $\mathcal{L}_\text{d} = \sum_{i,j}\omega_i\omega_j|z_i-z_j|$, where $\omega_i = \alpha_i\,\mathcal{G}_i(\mathbf{u}(\mathbf{x}))\prod_{j=1}^{i-1} (1 - \alpha_j\,\mathcal{G}_j(\mathbf{u}(\mathbf{x})))$ is the blending weight of $i-$th intersection and $z_i$ is the intersection depth. Since 2DGS explicitly models the normals, we can align the rendered normals $\mathbf{n}_i$ with the normals $\mathbf{N}$ computed from depth via consistency loss: $\mathcal{L}_\text{n} = \sum_{i} \omega_i (1-\mathbf{n}_i^\top\mathbf{N})$.
The regularization term for the ray $\mathbf{u}(\mathbf{x})$ is: $\mathcal{L}_\text{Reg} = \gamma_\text{d} \mathcal{L}_\text{d} + \gamma_\text{n} \mathcal {L}_\text{n}$, and we set $\gamma_\text{d} \!=\! 1000$ and $\gamma_\text{n} \!=\! 0.2$ following LaRa~\cite{chen2025lara}. Our total loss function is defined as:
\begin{align}
\mathcal{L} = \mathcal{L}_{\text{Image}} + \mathcal{L}_{\text{Reg}} + \mathcal{L}_{\text{Geometry}}.
\label{eqn:loss_total}
\end{align}

In practice, training a multi-channel model directly can be challenging, especially due to the limited availability of high-quality training data for roughness and metallic components compared to albedo. To address this, we implement a two-stage training strategy. In the first stage, we freeze the roughness and metallic components of the material Gaussian, focusing exclusively on training the albedo component. In the second stage, we include all material components in the training process. 
This approach improves the training quality and leads to faster convergence. 

\subsection{Relighting}
\label{sec:Relightable Gaussian}
In our framework, we leverage the classic rendering equation to formulate the outgoing radiance of a surface point $\bm{x}$ with normal $\bm{n}$ under varying lighting conditions:
\begin{equation}
\small
L_o(\bm{x}, \bm{v}) = \int_\Omega L_i(\bm{x}, \bm{l}) f_r(\bm{l}, \bm{v}) (\bm{l} \cdot \bm{n}) d\bm{l},
\label{eq:rendering}
\end{equation}
where $\Omega$ represents the upper hemisphere centered at $\bm{x}$, $\bm{l}$ and $\bm{v}$ denote incident and view directions respectively. $L_i(\bm{x}, \bm{l})$ denotes the radiance received at $\bm{x}$ from $\bm{l}$. We follow Cook-Torrance microfacet model~\cite{cook1982reflectance, walter2007microfacet} to formulate the bidirectional reflectance distribution function (BRDF) $f_r$ as a function of albedo $\bm{a} \in [0, 1]^3$, metallic $m \in [0, 1]$, and roughness $\rho \in [0, 1]$:
\begin{equation}
\small
f_r(\bm{l}, \bm{v}) =
\underbrace{(1 - m) \frac{\bm{a}}{\pi}}_\text{diffuse} +
\underbrace{
\frac{
    DFG
}{
    4 (\bm{n} \cdot \bm{l}) (\bm{n} \cdot \bm{v})
} 
}_\text{specular},
\label{eq:brdf}
\end{equation}
where microfacet distribution function $D$, Fresnel reflection $F$, and geometric shadowing factor $G$ are related to the surface roughness $\rho$. With our material Gaussians that store all the PBR properties and given the view direction $\bm{l}$, we can render the albedo $\bm{a}$, roughness $\bm \rho$, metallic $\bm m$ and normal map $\bm{n}$ with 2DGS rasterization, then we can apply different environment light maps to the BRDF function $f_r(\cdot)$ to acquire the relighting results.

%% file: sections/5-experiments.tex
\section{Experiments}

\begin{figure*}
    \centering
    \includegraphics[width=1.01\textwidth]{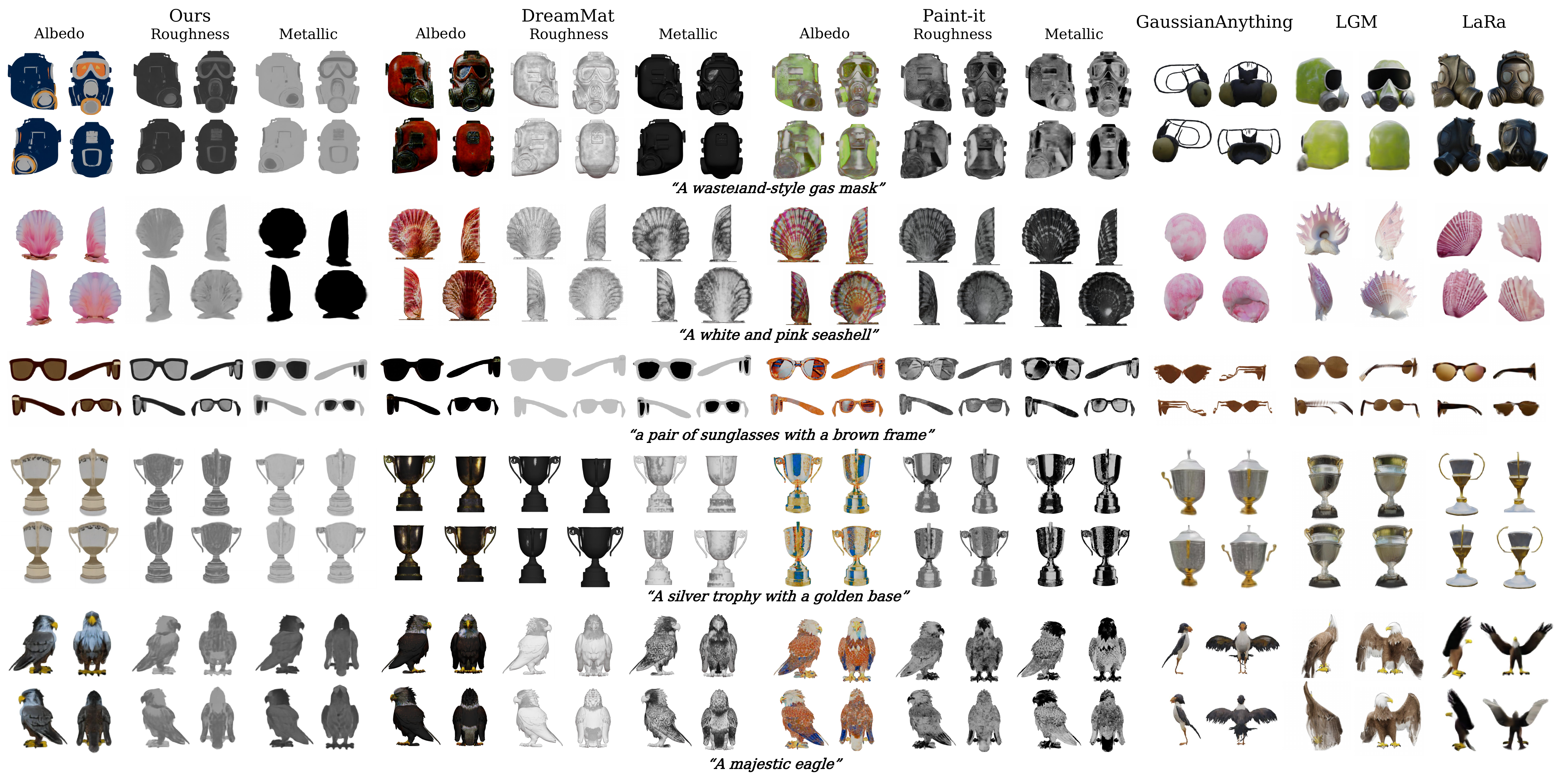}
    \caption{
    \textbf{Qualitative comparison.} 
    Given various text prompts, our method generates 3D assets with material properties, and achieves better geometry and appearance quality comparing to other Gaussian-based generative methods.
    }
    \vspace{2mm}
    \label{fig:main-comparison}
\end{figure*}

\subsection{Implementation Details}
In our multiview PBR model, we train two sub-models for albedo and roughness/metallic, initialized from MVDream~\cite{shi2023mvdream}. For the albedo model, we train the first two blocks in the input blocks and the last one block in the output blocks. For roughness/metallic model, we inflate the trainable parts mentioned in albedo model and train the full parameters due to the larger gap between roughness/metallic and shaded images. We use depth and normal ControlNets~\cite{zhang2023adding} for both models. 
The multiview PBR dataset is rendered from objects in the Objaverse~\cite{deitke2023objaverse} dataset with Blender, which contains about 80,000 entries, each with a descriptive caption sourced from Cap3D~\cite{luo2024scalable}.

In our material Gaussian reconstruction model, the input images are all at a resolution of $512 \times 512$. After DINO encoding, the shape of the image feature is $768 \times 32 \times 32$, and the resolution of the embedding volume is $32$ with $256$ channels. Following LaRa, the volume decoder consists of 12 group attention layers with $G \!=\! 16$ groups, producing a Gaussian volume of size $64 \times 64 \times 64 \times 80$. We choose $K \!=\! 2$ primitives for each voxel and constrain the offset radius to $r=1/32$ of the length of the bounding box. The model is trained for a total of 80 epochs, with the first and second stages each accounting for half of the total training duration. We train the model on G-buffer Objaverse (Gobjaverse)~\cite{qiu2024richdreamer} dataset and filter out the `low-quality' objects as defined in the dataset.

\begin{table*} 
    \begin{center}
    \setlength{\tabcolsep}{16pt}
    \caption{\textbf{Quantitative comparison.} All scores are mean value across all samples. Here GA is GaussianAnything~\cite{lan2024gaussiananything}.}
    % \resizebox{0.48\textwidth}{!}{
    {
    \begin{tabular}{{l}c*{9}{c}}
        \toprule
        \makecell{Metrics / Methods} & GA & LaRa & LGM & Paint-it & DreamMat & \textbf{Ours}\\
        \midrule
        Geometry CLIP $\uparrow$     & 26.66  & 27.02 & 27.84  & - & -  & \textbf{29.87} \\
        Appearance CLIP $\uparrow$   & 28.28  & 28.77 & 29.31  & 27.65  & 28.49 & \textbf{30.48} \\
        FID  $\downarrow$            & 121.49 & 97.95 & 101.60 & 113.87 & 105.32 & \textbf{89.55} \\
        Time    $\downarrow$         & - & - & - & 40min & 1h15min  & \textbf{30sec} \\
        \bottomrule
    \end{tabular}
    }
    % }
    \label{tab:geo_app_clip}
\end{center}
\end{table*}

\subsection{Training Datasets}
For our multiview PBR diffusion, we use images rendered from objects in the Objaverse~\cite{deitke2023objaverse} dataset with Blender. We filter out objects without material maps. For roughness and metallic maps, we select four pairs of roughness and metallic rendered images with elevations of 0 and azimuths of 0, 90, 180, and 270 degrees to calculate the CLIP score~\cite{radford2021learning}. Objects with scores exceeding 0.95 are filtered out, as they have almost identical roughness and metallic maps, suggesting low-quality material maps. Following the camera settings in MVDream~\cite{shi2023mvdream}, we render each object from 32 views at a fixed elevation angle and camera distance in each setting, and sample three times on elevation and camera distance. The rendered image sets include albedo, roughness, metallic, normal, and depth map for each view. For normal rendering, we disable the normal UV map with fine details that comes with the model, considering that normals are generally calculated from the model's geometry and do not exhibit such fine details during inference. The filtered subset contains about 80,000 entries, each with a text prompt captioned by Cap3D~\cite{luo2024scalable}.

For our material Gaussian reconstruction model, we use G-buffer Objaverse (Gobjaverse)~\cite{qiu2024richdreamer} dataset, which is based on Objaverse~\cite{deitke2023objaverse}. This multiview rendering dataset includes 280,000 scenes with PBR data. After removing blank albedo maps and 'low-quality' objects as defined in the dataset, we obtain a subset of 100,000 albedo maps. For roughness and metallic maps, we notice that many images are single-colored, likely due to default settings in Gobjaverse for missing channels. To address this, we implement a simple filtering strategy to exclude images where the roughness or metallic foreground pixel value is unique, resulting in a subset of 60,000 images.

\begin{figure*}
    \centering
    \includegraphics[width=0.75\textwidth]{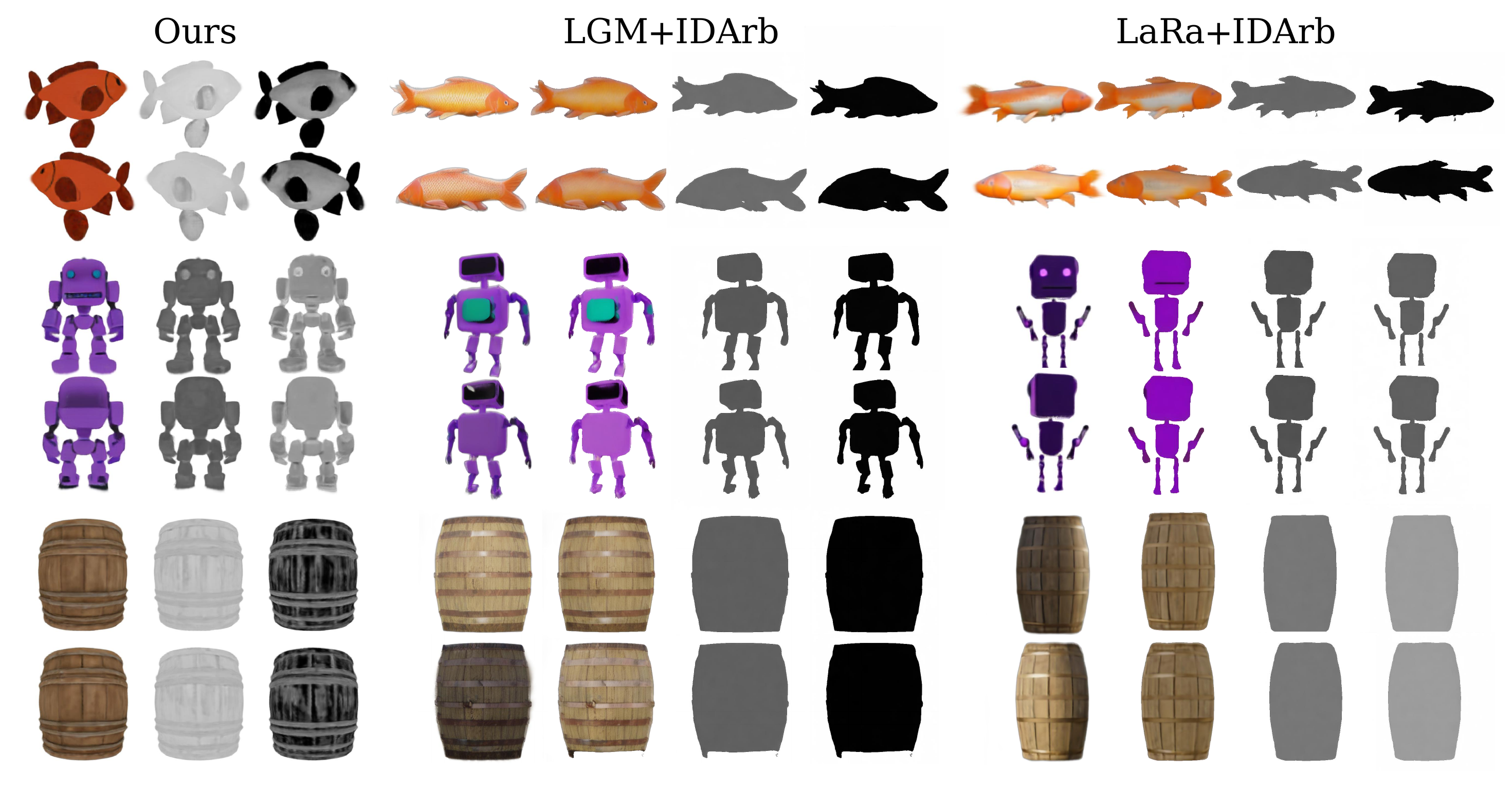}
    \caption{
    \textbf{Comparison with post-processing for material estimation}. For Ours, the first, second, and third columns are albedo, roughness, and metallic, respectively. For LGM~\cite{tang2025lgm} and LaRa~\cite{chen2025lara}, the first column is the shaded RGB images, and the last three columns are decomposed albedo/roughness/metallic by IDArb~\cite{li2024idarb}, 
    denote as \emph{LGM+IDArb} and \emph{LaRa+IDArb}.
    The text prompts are \emph{"An orange and red carp"}, \emph{"A cartoon-style purple robot"}, and \emph{"A wooden barrel"}, respectively.
    }
    \label{fig:idarb_comparison}
\end{figure*}

\subsection{Details on Proxy Geometry Generation}

Since our multiview PBR diffusion is conditioned on the multiview depth and normal maps as geometry prior, we need to generate them from text input. 
Initially, we plan to use the multiview depth and normal diffusion provided by Richdreamer~\cite{qiu2024richdreamer} to obtain four multiview depth and normal maps directly from the input text. However,  upon testing, we find that this model directly resize the image to $32\times32$ as the latent feature and dose not train the VAE with depth and normal additionally, resulting in generated images that are very blurry and noisy. As a result, we shift our approach to generate a mesh from the text prompt and then rendering the mesh.
We first input the text into Stable Diffusion 2.1~\cite{rombach2022high} to obtain the corresponding image, and then remove the background of the image and import it into CraftsMan~\cite{li2024craftsman} to create a 3D model. The resulting mesh goes through a round of redundant faces deletion, 
and then we normalize it to fit within a bounding box of $[-0.5, 0.5]$ and translate it to the origin.
Finally we render the mesh with Pytorch3D~\cite{ravi2020pytorch3d} to obtain multi-view depth and normal maps, the camera distance is set to $1.5$.

\subsection{Coarse-to-Fine Decoding of Reconstruction Model}
\label{sec:gs_para_decoder}
Following LaRa~\cite{chen2025lara}, we extract 2D Gaussian primitive shape and appearance parameters from the Gaussian volume using a coarse-to-fine decoding process. In the \emph{coarse} decoding stage, Gaussian volume features are fed into a lightweight MLP, which outputs a set of $K$ Gaussian parameters for each voxel. In order to better preserve the appearance, a \emph{fine} decoding stage is added to guide the fine texture prediction. Specifically, the primitive centers $p^k_i$ are projected onto the coarse renderings, \ie, RGB image $\hat{I}$, depth image $\hat{D}$, and alpha map $\hat{A}$), to incorporate the coarse renderings for each primitive using the camera poses $\pi$,
\begin{align}
\mathcal{X}_{p^k_i} = \left(I_{p^k_i},\hat{I}_{p^k_i}, \hat{D}_{p^k_i}, \hat{\mathrm{A}}_{p^k_i}\right) = \Phi \left(\mathcal{P} \left(p^k_i, pi \right), \oplus \left[I,\hat{I}, \hat{D}, \hat{\mathrm{A}}\right] \right)
\end{align}
where $\mathcal{P}$ denotes the point projection, $\oplus$ is a concatenation operation along the channel dimension, and $\Phi$ represents bilinear interpolation. 
We then implement a point-based cross-attention layer between the features of point $\mathcal{X}_{p^k_i}$ and the primitive voxel. The results are subsequently fed into an MLP, which is designed to predict the residual spherical harmonics:
\begin{align}
\text{SH}_{i,k}^\textit{residuals} &= \text{MLP}\left(\text{CrossAttn}\left(\mathcal{X}_{p^k_i}, V^i_{\mathcal{G}} \right) \right) \text{,} \\
\text{SH}_{i,k}^\textit{fine} &= \text{SH}_{i,k}^\textit{coarse} + \text{SH}_{i,k}^\textit{residuals} 
\end{align}
Finally, our work takes advantage of 2DGS~\cite{huang20242d} to enable efficient high-resolution rendering, and both coarse and fine stages are differentiable and updated simultaneously.

% \begin{table*}[!t]
\begin{table} \centering
\begin{center}
\setlength{\tabcolsep}{3pt}
\caption{
\textbf{User study.} The rating is of scale 1-5, where the higher the better.
}
\resizebox{0.48\textwidth}{!}{
\begin{tabular}{lcccc}
\toprule 
\makecell[l]{\textbf{Methods}} & \makecell{\textbf{Multiview} \\ \textbf{Consistency}} $\uparrow$ & \makecell{\textbf{Textual} \\ \textbf{Alignment}} $\uparrow$ & \makecell{\textbf{Geometric} \\ \textbf{Integrity}} $\uparrow$ & \makecell{\textbf{Overall} \\ \textbf{Quality}} $\uparrow$ \\
\midrule
GA~\cite{lan2024gaussiananything}  & 2.85 & 3.30 & 3.15 & 2.96 \\
LGM~\cite{tang2025lgm}                           & 3.38 & 3.74 & 3.51 & 3.63 \\
LaRa~\cite{chen2025lara}                         & 3.47 & 3.62 & 3.65 & 3.45 \\
Paint-it~\cite{youwang2024paint}                 & 3.06 & 3.30 & - & 3.27 \\
DreamMat~\cite{zhang2024dreammat}                & 3.15 & 3.55 & - & 3.40 \\
\textbf{MGM (Ours)}                              & \textbf{3.66} & \textbf{3.78} & \textbf{3.92} & \textbf{3.80}  \\
\bottomrule

\end{tabular}
}
\label{tab:comparison_userstudy}
\end{center}
\end{table}

\subsection{Baselines and Evaluation Metrics}
We conduct comparisons primarily with Gaussian Splatting based 3D generation methods and other PBR texture generation methods, including LaRa~\cite{chen2025lara}, LGM~\cite{tang2025lgm}, GaussianAnything~\cite{lan2024gaussiananything}, DreamMat~\cite{zhang2024dreammat} and Paint-it~\cite{youwang2024paint}. LaRa and LGM are two methods that follow LRM~\cite{hong2023lrm} fashion with Gaussian Splatting representation. GaussianAnything is a native 3D diffusion model that employs a 3D VAE to decode and generate 2D Gaussians. It's important to note that these methods are limited to generate shaded textures rather than PBR materials. All the 3D Gaussians based methods achieve fast generation, requiring less than a minute to produce one scene. 
DreamMat and Paint-it are two PBR texture generation methods based on 2D-lifting optimization, we use the geometry generated by CraftsMan~\cite{li2024craftsman} as input for these two methods.

Evaluating the quality of text-to-3D models is challenging due to the lack of standard metrics. 
To objectively evaluate the quality, we adopt the evaluation method from RichDreamer~\cite{qiu2024richdreamer} and use the CLIP model (ViT-B-32) to calculate the \textit{Geometry CLIP Score} and \textit{Appearance CLIP Score}. The CLIP score ranges from 0 to 1 and is multiply by 100. 
We also calculated FID~\cite{heusel2017gans} to evaluate the generated appearance quality, as well as the time required for the generation to measure the efficiency of different methods. 
We select 148 text prompts from various 3D generation works~\cite{lan2024gaussiananything, su2024gt23d, xiang2024structured, siddiqui2024meta, tang2025lgm}, render 24 predefined views for each object, and compute the average Geometry/Appearance CLIP score and FID. For our approach, we randomly select one of six ambient maps as the light source to synthesize the final rendering for evaluation, the light maps are shown in the Appendix.

Since DreamMat~\cite{zhang2024dreammat} and Paint-it~\cite{youwang2024paint} do not generate geometry, we do not calculate the geometry-related metrics of these two methods, \ie, Geometry CLIP score and Geometric Integrity. We also do not calculate the inference time of Gaussian-based generation methods, \ie, GaussianAnything~\cite{lan2024gaussiananything}, LGM~\cite{tang2025lgm} and LaRa~\cite{chen2025lara}, which are roughly comparable to our method.

\subsection{Main Comparison}
\noindent \textbf{Qualitative Results.} 
As shown in Figure~\ref{fig:main-comparison}, we render images from the generated 3D assets for visualization. The results produced by our method align appropriately with the input text and demonstrate strong generalization capabilities. For the geometry quality, other methods sometimes exhibit artifacts such as fragmentation or floating elements, whereas our method preserves a complete and detailed geometric topology. For the generated textures, other methods may produce blurry or unrealistic textures and the lighting factors cannot be removed. In contrast, our approach yields high-quality PBR materials with rich texture details, and the materials are closely aligned with the text inputs. Notably, the albedo maps generated by our approach do not contain any lighting or shadow effects, and the roughness and metallic maps accurately reflect the necessary details for different regions of the model.
In addition to the single-object results, we also find that the multi-object generated by our method exhibits superior geometry and texture appearance compared to the baseline methods. This improvement can be attributed to our incorporation of geometric prior guidance from multiview depth and normal maps in the inference stage of the multiview PBR diffusion. We include the results of multi-object generation in Appendix.

\noindent \textbf{Quantitative Results.} 
Table~\ref{tab:geo_app_clip} demonstrates that our method achieves state-of-the-art performances on geometry CLIP score, appearance CLIP score and FID compared to other methods. In particular, the geometry CLIP score is significantly better than other methods, indicating that our geometry prior is of great help in improving the geometric quality. Our method outperforms other methods in quantitative assessments and the results consistently align with human subjective judgments in qualitative evaluations. 
And our method can generate within one minute, while methods like DreamMat~\cite{zhang2024dreammat} and Paint-it~\cite{youwang2024paint} usually require more than dozens of minutes for optimization.  

\noindent \textbf{User Study.} 
To further assess the visual quality of the generated 3D models, we conduct a comprehensive user study with 12 volunteers. Each participant receives 8 examples, along with the corresponding text prompts. There are 4 evaluation indicators: overall quality, text consistency, multiview consistency, and geometric integrity, each scored from 1 to 5, with 1 being the worst and 5 being the best. Table~\ref{tab:comparison_userstudy} presents the results of our user study. In terms of multiview consistency, text consistency, geometric integrity, and overall quality, our method receives the highest scores.
We also conducted a user study on the quality of PBR generation in Table~\ref{tab:comparison_userstudy_pbr}, comparing with DreamMat~\cite{zhang2024dreammat}, Paint-it~\cite{youwang2024paint}, and LGM~\cite{tang2025lgm}/LaRa~\cite{chen2025lara} with IDArb~\cite{li2024idarb} post-processing. Our method achieved the highest scores in albedo, roughness, and metallic quality.

In the appendix, we further present several generation examples based on corresponding text prompts with all the baseline methods, and we add GVGEN~\cite{he2025gvgen} for comparison which is also a 3D diffusion model using a 3D U-Net to output 3D Gaussians.
And we show more of the generated material Gaussian with appropriate PBR properties, with the realistic rendering results under different physics world lighting. 
In addition to single-object generation, 3D generation also includes multi-object generation, we show the comparison results of multi-object generation in the appendix as well. It can be seen that under the guidance of geometric priors, the multiple objects we generate have a more complete and reasonable geometry, as well as rich texture colors and semantically aligned PBR materials. And we present the quantitative evaluation and the user study of multi-object generation, respectively. The superiority of our method over the single-object generation is more prominent obviously.

We also compared our multiview PBR diffusion with other multiview generation models, ie, MV-Adapter~\cite{huang2024mv} and MVDream~\cite{shi2023mvdream}, both of which can produce multiview images based on the text input. Some generation results are shown in the appendix. We can find that MV-Adapter and MVDream can only generate images with undecoupled light and shadow, while our multiview PBR diffusion generates PBR images without any light and shadow, and the roughness and metallic are highly consistent with the text prompts and have rich details. The results of the quantitative comparison of multiview generation are all given in the appendix.

\subsection{Comparison with Post-Processing on Material Estimation}
To demonstrate that our model can generate semantically accurate materials with high-quality details, we consider adopting post-processing modules like those in~\cite{hong2024supermat,li2024idarb,zeng2024rgb} based on other large reconstruction models that producing shaded images to extract materials. Here, we choose IDArb~\cite{li2024idarb} as it decomposes PBR materials from multiview shaded images with high quality. Figure~\ref{fig:idarb_comparison} shows our materials and the results of LGM and LaRa after PBR extraction by IDArb.
It is observed that the albedo generated through the post-processing method can partially eliminate light and shadow effects. However, the extracted roughness and metallic maps are monochromatic, lacking detail and exhibiting poor alignment with the text. In contrast, the PBR materials we generate not only produce an albedo free from lighting interference but also include highly detailed roughness and metallic maps that align closely with the textual semantics.

\begin{table} \centering
\begin{center}
\setlength{\tabcolsep}{6pt}
\caption{
\textbf{User Study on PBR.} The rating is of scale 1-5, where the higher the better.
}
\resizebox{0.47\textwidth}{!}{
\vspace{-1em}
\begin{tabular}{lcccc}
\toprule 
\makecell[l]{\textbf{Methods}} & \makecell{\textbf{Albedo} \\ \textbf{Quality}} $\uparrow$ & \makecell{\textbf{Roughness} \\ \textbf{Quality}} $\uparrow$ & \makecell{\textbf{Metallic} \\ \textbf{Quality}} $\uparrow$ \\
\midrule
LGM+IDArb~\cite{tang2025lgm,li2024idarb}       & 2.43 & 2.94 & 2.75 \\
LaRa+IDArb~\cite{chen2025lara,li2024idarb}     & 2.56 & 2.90 & 2.78 \\
Paint-it~\cite{youwang2024paint}               & 3.17 & 3.10 & 3.13  \\
DreamMat~\cite{zhang2024dreammat}              & 3.29 & 3.45 & 3.36  \\
\textbf{MGM (Ours)}                            & \textbf{3.97} & \textbf{3.55} & \textbf{3.60}  \\
\bottomrule
\end{tabular}
}
\label{tab:comparison_userstudy_pbr}
\end{center}
\end{table}

\begin{figure*}
% \begin{figure}
    \centering
    \includegraphics[width=0.7\textwidth]{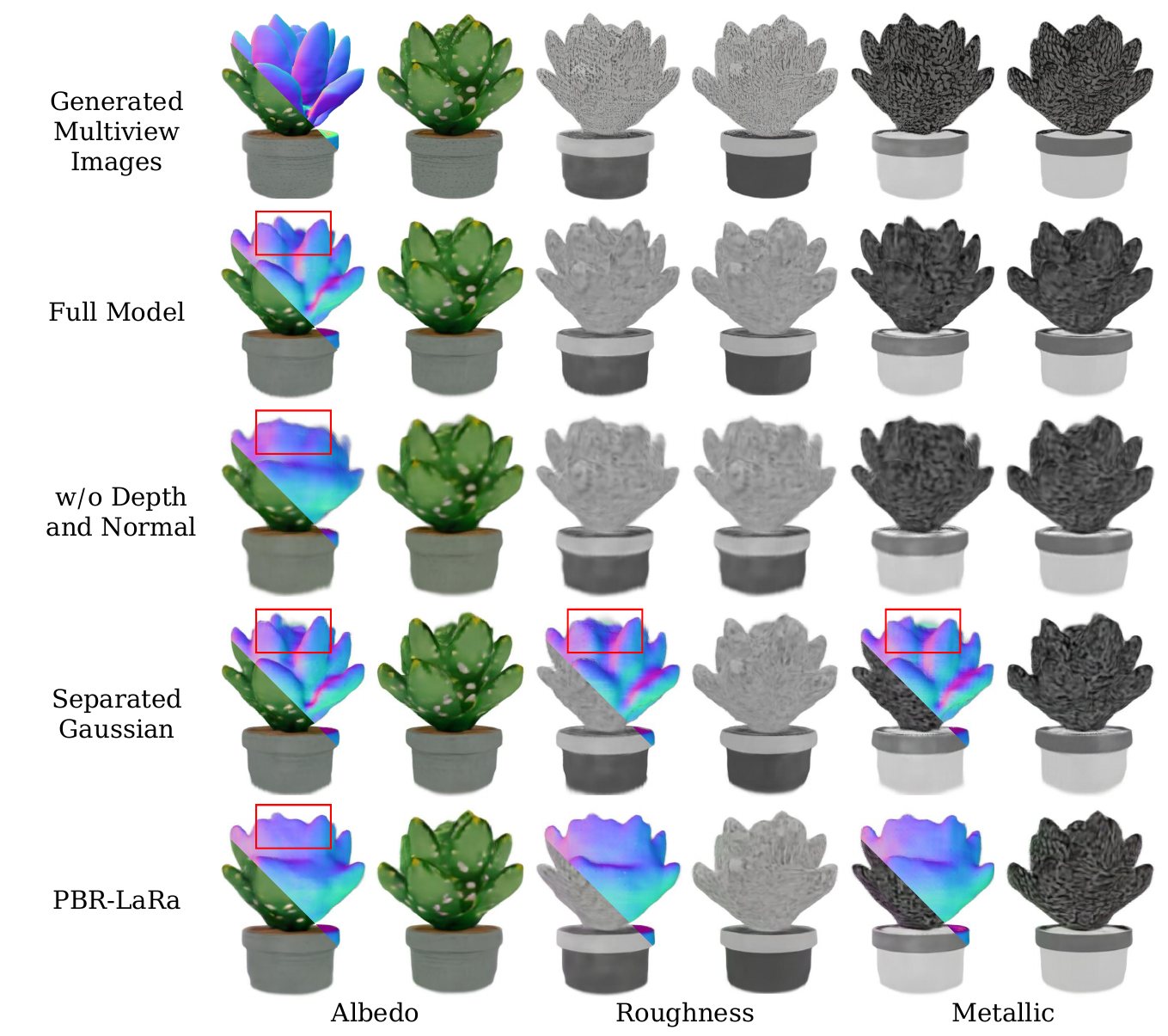}
    % \vspace{-1em}
    \caption{
    \textbf{Ablation study}. 
    The first row is the results generated by our multiview PBR diffusion. The next four lines are the reconstruction results corresponding to the ablation method. For the unified geometry model (full model, w/o depth and normal), we visualize the normal on albedo. For the reconstruction model with three independent channels, we visualize the normals on the PBR separately.
    }
    \label{fig:ablation_method}
\end{figure*}

\begin{table*}[t] \centering
    \caption{\textbf{Quantitative comparison on ablation study.} All scores were mean value across all samples, values the higher the better. Here \emph{SHs-degree=1} and \emph{SHs-degree=2} represent using first- and second-order spherical coefficients to model color in 2DGS, respectively.}
    {
    \begin{tabular}{lc*{9}{c}}
        \toprule
        \makecell{Metrics / Methods} & \textbf{Full Model} & w/o Depth and Normal & Seperated Models & PBR-LaRa & SHs-degree=1 & SHs-degree=2\\
        \midrule
        Geometry CLIP $\uparrow$    & \textbf{29.87} & 28.92 & 29.39 & 28.70 & 29.68 & 29.51\\
        Appearance CLIP $\uparrow$  & \textbf{30.48} & 29.85 & 29.64 & 29.22 & 30.14 & 30.05\\
        FID $\downarrow$            & \textbf{89.55} & 95.47 & 93.92 & 97.20 & 90.56 & 90.83 \\
        \bottomrule
    \end{tabular}
    }
    \label{tab:geo_app_clip_ablation}
    % \vspace{-2mm}
\end{table*}

\begin{figure}
    \centering
    \includegraphics[width=0.49\textwidth]{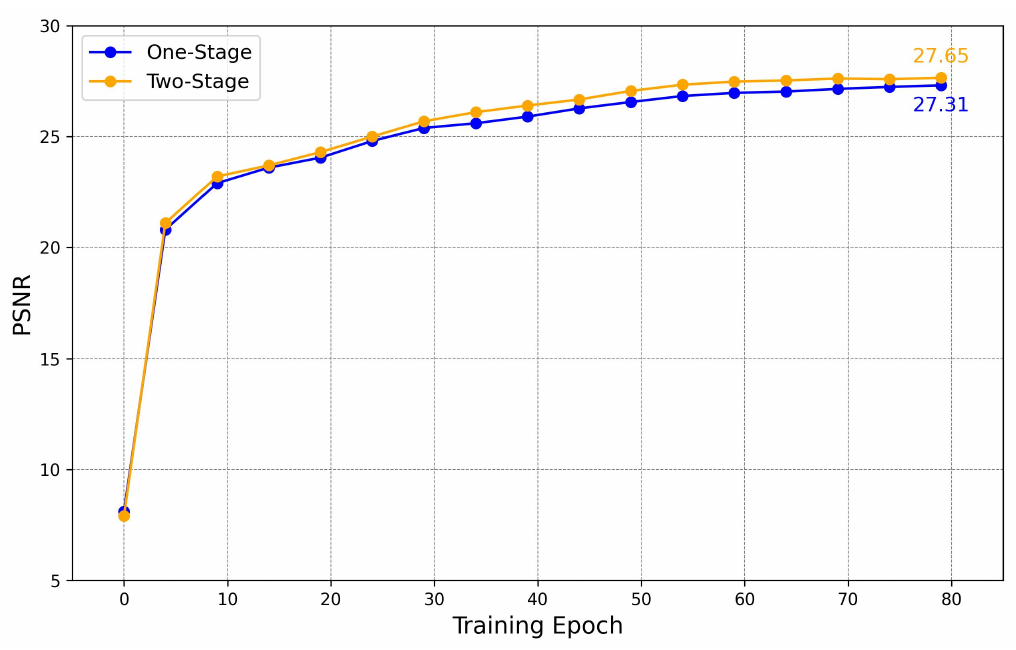}
    \caption{
    Training PSNR curves of one-stage and two-stage strategies respectively.
    }
    \label{fig:appendix_1_or_2_stage}
\end{figure}

\subsection{Ablation Study}
To validate the effectiveness of each component of our model, we conduct an ablation study on the text prompt \emph{"A green potted succulent plant"} to generate materials. The qualitative results are shown in Figure~\ref{fig:ablation_method}, and the quantitative results is presented in Table~\ref{tab:geo_app_clip_ablation}, it can be seen that the indicators deteriorate when our proposed components are disabled one by one.
Additionally, we train the models using the first-order and second-order spherical harmonic coefficients, which the metrics are also slightly lower, proving that PBR images without lighting information are more suitable for the simplest color modeling, \ie, directly using 3 color channels. 

\noindent \textbf{Disable Depth and Normal.}
Depth and normal maps provide crucial spatial information for the Gaussian points. When these images are not included, the quality of the reconstructed geometry significantly deteriorates, becoming smoother and losing a lot of details with slightly blurry textures.

\noindent \textbf{Three Independent Gaussian Models.}
Training PBR component separately does not yield a substantial difference in reconstruction quality compared to combined training. However, this approach results in three different geometries during inference, as highlighted by the red boxes in the fourth row of Figure~\ref{fig:ablation_method}. This geometric inconsistency will lead to a notable decline in relighting quality.

\noindent \textbf{Reconstruction with the Model Trained on Shaded Data.}
Since PBR images do not contain light and shadow clues, the reconstruction model pre-trained on shaded data fails to produce  satisfactory results. If we integrate our multiview PBR diffusion directly into LaRa, the quality declines significantly, as LaRa struggles to accurately predict the geometric details of the object due to the domain gap between these data distributions.

In Figure~\ref{fig:appendix_1_or_2_stage}, we also plot the training PSNR curves for both one-stage and two-stage training strategies proposed at the end of section~\ref{sec:Material Gaussian Reconstruction Model}, demonstrating that the two-stage strategy yields superior reconstruction quality and faster convergence.